\pdfoutput=1

\documentclass[11pt]{article}

\usepackage[final]{acl}

\usepackage{times}
\usepackage{latexsym}
\usepackage{booktabs}
\usepackage{multirow}
\usepackage{colortbl}
\usepackage{scalefnt}
\usepackage{subcaption}
\usepackage{enumitem}
\usepackage{algorithm}
\usepackage{algpseudocode}
\usepackage{amsmath}
\usepackage{setspace}
\usepackage{amsmath}
\usepackage{algorithm}
\usepackage{algpseudocode}
\usepackage{amsfonts}

\usepackage[T1]{fontenc}

\usepackage[utf8]{inputenc}

\usepackage{microtype}

\usepackage{inconsolata}
\usepackage{wrapfig}

\usepackage{graphicx}

\usepackage{varwidth}
\usepackage{tikz}
\usetikzlibrary{calc}
\usetikzlibrary{positioning, shapes, fit, backgrounds, decorations.pathreplacing}
\usepackage{fontawesome5}

\newcommand{\method}[0]{{LongGenBench}}
\NewDocumentCommand{\dom}{ mO{} }{\textcolor{blue}{\textsuperscript{\textit{Dom}}{#1}}}

\NewDocumentCommand\emojicheck{}{\includegraphics[scale=0.7]{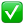}}

\NewDocumentCommand\emojicross{}{\includegraphics[scale=0.7]{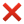}}

\title{\textsc{\method{}}: Long-context Generation Benchmark}

\author{
Xiang LIU $\qquad$ Peijie DONG $\qquad$ Xuming HU$^\dagger$ $\qquad$ Xiaowen CHU$^\dagger$ \\
The Hong Kong University of Science and Technology(Guangzhou)\\
 \texttt{\{xliu886,pdong212\}@connect.hkust-gz.edu.cn} \\
 \texttt{xuminghu@hkust-gz.edu.cn} \quad \texttt{xwchu@ust.hk}
  }

\begin{document}

\maketitle
\begin{abstract}
Current long-context benchmarks primarily focus on retrieval-based tests, requiring Large Language Models (LLMs) to locate specific information within extensive input contexts, such as the needle-in-a-haystack (NIAH) benchmark. Long-context generation refers to the ability of a language model to generate coherent and contextually accurate text that spans across lengthy passages or documents. While recent studies show strong performance on NIAH and other retrieval-based long-context benchmarks, there is a significant lack of benchmarks for evaluating long-context generation capabilities. To bridge this gap and offer a comprehensive assessment, we introduce a synthetic benchmark, \textbf{\method{}}, which is designed to evaluate the long-context generation capabilities of large language models (LLMs), with a particular focus on consistency in logical flow. \method{}  redesigning the format of questions and necessitating that LLMs respond with a single, cohesive long-context answer. Upon extensive evaluation using \method{}, we observe that: (1) both API accessed and open source models exhibit performance degradation in long-context generation scenarios, ranging from 1.2\% to 47.1\%; (2) different series of LLMs exhibit varying trends of performance degradation, with the \textsc{Gemini-1.5-Flash} model showing the least degradation among API accessed models, and the \textsc{Qwen2} series exhibiting the least degradation in \method{} among open source models. The code is available at \url{https://github.com/Dominic789654/LongGenBench}.

\end{abstract}

\section{Introduction}
\begin{figure}[!h]
\vskip 0.2in
\begin{center}
\includegraphics[width=1\linewidth,trim=0 0 0 0]{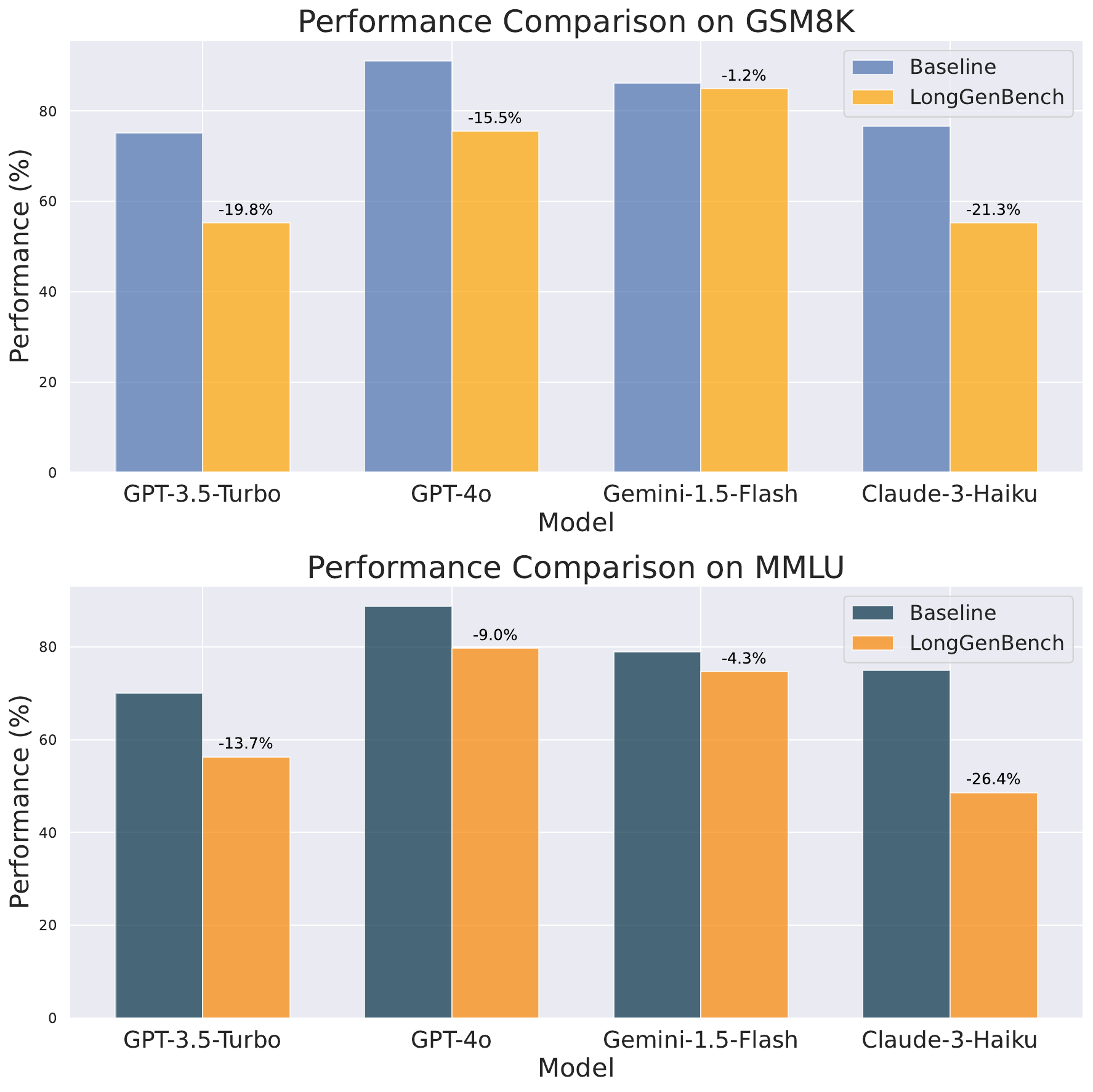}
\caption{Performance Comparison of LLMs on GSM8K and MMLU datasets using \method{} to assess their long-context generation capabilities. It is observed that mainstream LLMs exhibit performance degradation when tasked with long-context generation.}
\label{fig:performance_conparison}
\end{center}
\end{figure}
\def\thefootnote{$\dagger$}\footnotetext{Corresponding Author.}

Large Language Models (LLMs) have become pivotal in tackling NLP downstream tasks such as summarization and question answering that require interpreting extensive context from books, reports, and documents, sometimes spanning tens of thousands of tokens~\citep{raffel2020exploring, brown2020language, chowdhery2022palm, tay2022unifying, touvron2023llama2, tang2023fusionai, zhang2023dissectingruntimeperformancetraining}. Recent advances in long-context technology in ML system field~\citep{flash-attn,flash-attn2,jacobs2023deepspeed,xiao2024efficient} and model architecture design~\citep{chen2023extending, xiong2023effective,chen2023longlora,peng2024yarn, Pruner-Zero} have significantly improved the ability of LLMs to process increasingly large input context lengths~\citep{liu2024world,young2024yi}, such as Gemini-1.5-pro model can handle the 1,500-page document~\citep{geminiteam2024gemini}. Although previous studies~\citep{jamba, grok,geminiteam2024gemini,claude3,deepseekv2} often employ synthetic tasks like passkey retrieval~\citep{mohtashami2023landmark} and needle-in-a-haystack (NIAH)~\citep{needle} to evaluate the long-context capability of these LLMs, such tasks primarily test retrieval skills and do not fully assess other aspects of long-context generation. 

Long-context generation refers to the ability of a language model to generate coherent and contextually accurate text that spans across a lengthy passage or document. This capability involves maintaining the thematic continuity, logical flow, and consistency of details over extended sequences of text, which can include multiple paragraphs, pages, or even entire documents.

To facilitate further research in this area, we propose the \textbf{Long}-context \textbf{Gen}eration \textbf{bench}mark (\textbf{\method{}}), a new benchmark specifically designed to evaluate the long-context generation capabilities of LLMs, with a particular focus on consistency in logical flow. \method{} synthesizes a dataset from current popular LLM benchmarks, redesigns the input format, and includes multiple questions within a single query. The \method{} requires LLMs to generate a comprehensive long-context response that sequentially addresses each question. To achieve better performance in \method{}, LLMs need to maintain consistency regardless of whether the previous generation part is correct or incorrect. In \method{}, evaluating the quality of these long-context responses is straightforward: simply compare the generated answers with the ground truth.

Our study evaluates the performance of various language models using the \method{} approach across different datasets, specifically \textit{\method{}-MMLU}, \textit{\method{}-GSM8K}, and \textit{\method{}-CSQA}. Figure \ref{fig:performance_conparison} displays the performance of four powerful API accessed models tested in both the baseline scenario, which involves single-answer generation, and the \method{} scenario, which focuses on long-context generation. It is notable that the Gemini-1.5-Flash model exhibits the lowest performance degradation in long-context generation tasks, surpassing the GPT-4o. Additionally, we conducted analyses on open-source models, revealing a general correlation between baseline performance and \method{} performance. \textbf{Models with higher baseline scores tend to show smaller declines in long-context generation tasks.} Models like Qwen2-72B-Instruct and DeepSeek-v2-Chat, both with high baseline scores, also exhibit minimal performance degradation. However, there are exceptions, such as LLaMA-3-70B-Instruct, which, despite its high baseline performance, experiences significant performance drops. Moreover, \textbf{model size influences performance}, as larger models within the same series, such as the LLaMA-3 and Qwen2 series, demonstrate smaller declines. \textbf{Different architectures show varying trends} in performance degradation; for example, LLaMA-3-8B-Instruct shows a performance degradation of 47.1\% on GSM8K, while ChatGLM4-9B-Chat only experiences a 10.8\% drop, despite having similar baseline performances. \textbf{Consistency across tasks} is observed, with models like LLaMA-3-8B-Instruct consistently showing significant drops on all datasets, whereas models such as Qwen2-72B-Instruct and DeepSeek-v2-Chat maintain minimal declines across all datasets, underscoring their resilience in long-context generation tasks. These findings highlight the varying capabilities of different models to maintain accuracy over extended text generation and provide valuable insights for future model development and optimization.

Our contributions are as follows:
\begin{itemize}[leftmargin=*]
\vspace{-0.5em}
    \item We introduce \method{}, an effective approach for evaluating the long-context generation capabilities of language models across multiple datasets.
    \item We provide a comprehensive performance comparison between API accessed and open source models under the \method{} framework, revealing insights into how different models handle long-context generation tasks.
    \item Our analysis uncovers critical relationships in long-context generation tasks, including the correlation between baseline performance and \method{} performance, the impact of model size on performance decline, and the variation among different model architectures. Our detailed experiments establish consistent trends in performance degradation across different \method{} tasks, highlighting the importance of model resilience in long-context generation.
\end{itemize}

\section{Related Work}
\subsection{Long-context Language Models}

\begin{figure*}[!h]
\vskip 0.2in
\begin{center}
\includegraphics[width=1\textwidth,trim=0 0 20 0]{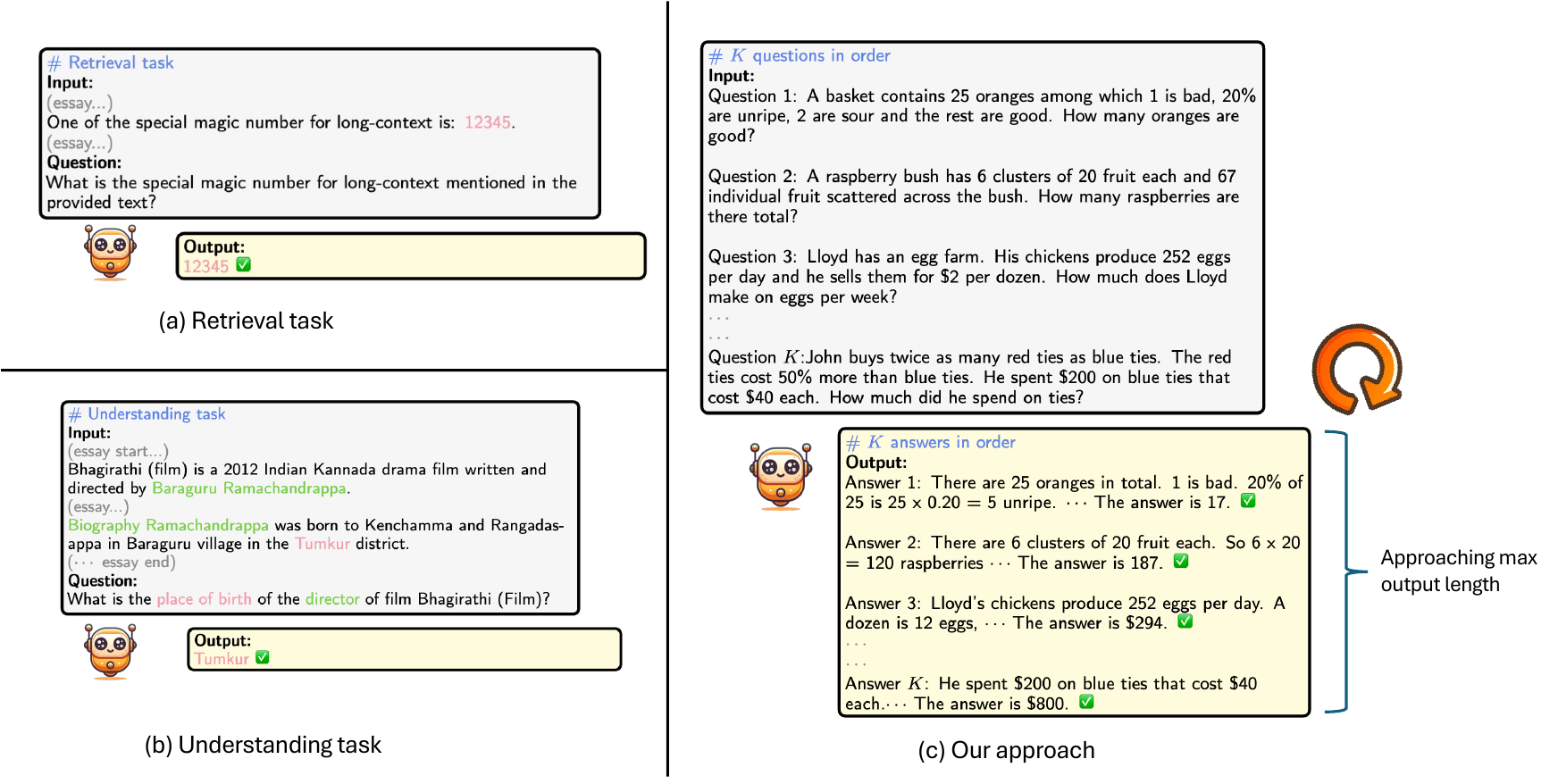}
\caption{Illustrations of previous long-context benchmarks and our proposed approach.
\textbf{(a) Retrieval task}: requires LLMs to retrieve the magic information hidden within an unrelated long context.
\textbf{(b) Understanding task}: requires LLMs to comprehensively understand a long essay and answer the specific question.
\textbf{(c) Our approach}: reconstructs the format of the dataset, requiring LLMs to sequentially understand and respond to each question in a single response. We run multiple iterations with different questions to evaluate the robustness of long-context generation capabilities. The length of the generated responses aims to approach the token limit.
}
\label{fig:main_fig}
\end{center}
\end{figure*}

Recent advancements in techniques such as efficient attention, long-term memory, extrapolative positional embedding, and context processing have spurred the development of numerous long-context LLMs \citep{huang_advancing_2023}. 
Efficient attention mechanisms like Flash attention \citep{flash-attn,flash-attn2} and Ring attention \citep{liu2023ring} have dramatically reduced memory demands for processing extensive contexts. Moreover, sparse attention methods, including shifted sparse attention in LongLoRA \citep{chen2023longlora}, dilated attention \citep{ding2023longnet}, and attention sinks \citep{han2023lm, xiao2024efficient}, further enhance long-context capabilities. 
For long-term memory, efficiency is achieved by caching previous contexts using recurrent mechanisms \citep{zhang2024soaring,bulatov2023scaling,martins2022former,wu2022memformer,mohtashami2023landmark}. 
Techniques for extrapolative positional embedding include ALiBi \citep{press2021alibi}, xPOS \citep{sun2023length}, and RoPE \citep{su2024rope}, along with their variants \citep{chen2023extending, xiong2023effective, peng2024yarn, liu20242, ding2024longrope, zhu2023pose}. 
In terms of context processing, key information is retained through retrieval augmentation \citep{xu2023retrieval, wang2023augmenting, tworkowski2024focused} and prompt compression \citep{jiang2023longllmlingua}. Innovative architectural designs such as state-space models \citep{gu2021s4,fu2022hungry,poli2023hyena,fu2023simple,gu2023mamba,dao2024transformers} and RWKV \citep{peng2023rwkv} are also being developed to effectively manage long-context inputs.

\subsection{Evaluation for Long-context Language Models}

Numerous investigations into long-context model benchmarks have primarily focused on retrieval and understanding tasks. 
In the realm of retrieval benchmarks, the datasets used are predominantly synthetic, enabling precise control over experimental conditions, such as input token length, and minimizing the influence of varied parametric knowledge from different training strategies. Recent research has extensively focused on synthetic tasks designed for retrieval \citep{needle, mohtashami2023landmark, longchat, liu2024lost,hsieh2024ruler,hu2024towards, hu-etal-2024-evaluating,zhang2024needlanguagespecificfactcheckingmodels}, with additional studies exploring the use of long contexts for various types of reasoning~\citep{tay2020long}.
For understanding benchmarks, LongBench~\citep{bai2023longbench} includes evaluations in a bilingual context, covering long-document question answering, summarization, and code completion tasks. ZeroSCROLLS~\citep{shaham2023zeroscrolls} and L-Eval~\citep{an2023eval} assess a wide array of realistic natural language tasks, such as long-document question answering and query-driven summarization. $\infty$-Bench~\citep{zhang2024infty} offers challenges that involve content spanning more than 100,000 tokens.

\section{\method{}}
We propose \method{}, a synthetic benchmark that is an efficient, low-cost approach focused on evaluating long-context generation in LLMs.

\subsection{Motivation}
Traditionally, evaluating LLMs for long-context scenarios involves inputting lengthy essays into the models, followed by either retrieval or comprehension questions as depicted in Figure~\ref{fig:main_fig}(a) and (b). The token length of these essays typically ranges from \{4K, 8K, 16K, 32K, 64K, 128K\}, with advanced long-context LLMs like Gemini-1.5~\citep{geminiteam2024gemini} being tested up to 1M tokens. However, these benchmarks tend to focus predominantly on the prompt tokens or input content, often neglecting the completion tokens or output content and the evaluation of performance regarding these aspects. Furthermore, traditional long-context benchmarks such as the NIAH test are costly, with a 128K NIAH test consuming 8M tokens.


\begin{algorithm}[!h]
\caption{Pipeline of \textbf{\method{}} } 
\label{alg:method_algorithm}
\begin{algorithmic}[1]
\Require System Prompt $S$, Questions $Q$, Number of Questions $K$, Number of Iterations $T$, Language Model $LLM$
\Ensure Long-Context Responses $R$
\State $R \gets \emptyset$
\For{$t \gets 0$ to $T-1$}
    \State $Q_t \gets Q[t \times K : (t+1) \times K]$
    \State $InputPrompt \gets S + \text{concatenate}(Q_t)$
    \State $Response \gets LM.\text{gen}(InputPrompt)$
    \State $ParsedResponse \gets \text{parse}(Response)$
    \State $R \gets R \cup \{ParsedResponse\}$
    \State $\text{verify}(ParsedResponse, Q_t)$
\EndFor
\State \Return $R$
\end{algorithmic}
\end{algorithm}

\subsection{Problem Definition}
In \method{}, the initial step involves redesigning the input prompt format to enable LLMs to generate long-context responses as illustrated in Figure~\ref{fig:main_fig}(c). We refine the system prompt and restructure the question format so that $K$ questions are sequentially concatenated after the system prompt. Subsequently, the LLMs are expected to adhere to this redesigned prompt and produce a coherent long-context response that answers all $K$ questions. These responses are then parsed to verify the answers to the $K$ questions, where the LLMs must maintain both the sequence and accuracy to demonstrate improved performance in \method{}. This process is repeated for $T$ iterations to assess the robustness of the LLMs' long-context generation capabilities at each length, with each iteration featuring unique questions.

The Algorithm \ref{alg:method_algorithm} gives a pseudocode outline for the \method{}.
The system prompt $S$ contains instructional information, while $Q$ is a list of questions from the original dataset. For each iteration $t$, a batch of $K$ questions, $Q_t$, is selected from $Q$ within the range $[t \times K : (t+1) \times K]$. The selected questions are concatenated to the system $S$ to form the $InputPrompt$. The language model $LLM$ generates a long-context response for the given $InputPrompt$. The response is added to the response set $R$, then parsed and verified for correctness and sequence. This process is repeated for $T$ iterations, with each iteration featuring a unique set of questions. The final output is the set of long-context responses $R$.

During the generation process, LLMs may accumulate correct or incorrect reasoning steps, which fall within the scope of \method{}'s evaluation. These models might generate errors during a single long-context session, and earlier mistakes can influence subsequent outputs. Assessing a model's performance in generating long texts involves evaluating how effectively it manages and mitigates these accumulated errors, and maintains consistency in logical flow. \method{} addresses this challenge by requiring models to handle and correct the impact of previous mistakes within a single long-context generation.

The conditional probability that the LLM generates the next token, given the prompt and the previously generated outputs, can be represented as:
\[ P(x_{i+1} \mid \text{\textit{InputPrompt}}, x_1, x_2, \ldots, x_i  )\]
Where $x_1, x_2, \ldots, x_i$ are the tokens generated in \method{}, the LLMs are required to produce the output based on the \textit{InputPrompt} and all previously generated tokens.

\subsection{Dataset Construction}
\method{} synthesizes three datasets from different domains: \textit{World Knowledge} from MMLU~\cite{hendrycks2020measuring}, \textit{Arithmetic} from GSM8K~\cite{cobbe2021training}, and \textit{Commonsense Reasoning} from CommonSenseQA~\cite{talmor2018commonsenseqa}. The MMLU dataset measures a model's ability to understand and reason across 57 diverse categories, using accuracy as the primary evaluation metric. The GSM8K dataset evaluates arithmetic problem-solving skills through 8,000 grade-school level math word problems, using the solving rate as the main metric. CommonSenseQA tests commonsense reasoning with multiple-choice questions based on ConceptNet, with accuracy as the evaluation metric. Appendix \ref{sec:app_format} provides details on how the synthesis process occurs.

\section{Expeirments Setting}
In this section, we describe the details of the baseline models and the \method{} approach, as well as their implementation in the subsequent experiments. All experiments were conducted three times, using the mean score to ensure robustness.

\begin{table}[!h]
\begin{sc}
\centering
\resizebox{1\linewidth}{!}{
\begin{tabular}{l|crcr}
\toprule
\multirow{2}{*}{Model} & \multirow{2}{*}{\begin{tabular}[c]{@{}c@{}}Access \\ Method\end{tabular}} & \multirow{2}{*}{\begin{tabular}[c]{@{}c@{}}Context \\ Length\end{tabular}} & \multirow{2}{*}{\begin{tabular}[c]{@{}c@{}}Max Output \\ Length\end{tabular}} & \multirow{2}{*}{\begin{tabular}[c]{@{}c@{}} Input\slash Output \\ Price\end{tabular}}  \\ 

&                                                                       
&
&
&\\ \midrule

GPT-3.5-Turbo          
& API 
& 16K
& 4K
& \$0.5 \slash \$1.5  \\

GPT-4o  
& API
& 128K
& 4K
& \$5 \slash \$15  \\

Gemini-1.5-Flash
& API
& 1024K
& 8K
& \$0.35 \slash  \$1.05 \\

Claude-3-Haiku  
& API
& 200K
& 4K 
& \$0.25 \slash \$ 1.25 \\ \midrule

LLaMA-3-8B
& Open Source
& 8K   
& -   
&  -   \\

LLaMA-3-70B
& Open Source
& 8K   
& -   
&  -   \\

Qwen2-7B
& Open Source
& 128K   
& -   
&  -   \\

Qwen2-57B
& Open Source
& 64K   
& -   
&  -   \\

Qwen2-72B
& Open Source
& 128K   
& -   
&  -   \\

ChatGLM4-9B
& Open Source
& 128K   
& -   
&  -   \\

DeepSeek-v2
& Open Source
& 128K   
& -   
&  -   \\

\bottomrule

\end{tabular}
}
\end{sc}
\caption{Comparison of context lengths for various LLMs.}
\label{tab:model_config}
\end{table}

\subsection{Models and Inference setup}
\label{sec:model_setup}
We evaluated multiple LLMs using \method{}, categorizing them into API accessed models and open-source models. For API accessed models, we selected GPT-3.5-Turbo~\citep{ouyang2022instructgpt, brown2020language}, GPT-4o~\citep{gpt-4o}, Gemini-1.5-Flash~\citep{geminiteam2024gemini}, and Claude-3-Haiku~\citep{claude3}. For open-source models, our selection included LLaMA-3-8B-Instruct, LLaMA-3-70B-Instruct~\citep{meta2024llama3}, Qwen2-7B-Instruct, Qwen2-57B-A14B-Instruct, Qwen2-72B-Instruct~\citep{qwen}, ChatGLM4-9B-Chat~\citep{zeng2022glm, du2022glm}, and DeepSeek-v2-Chat~\citep{deepseekv2}. The API accessed models are configured with a specific maximum output length, which constrains the number of output tokens due to the computational resources and commercial policies of each API provider. Table \ref{tab:model_config} provides detailed statistics for each model. For open-source models, we remove the \textsc{Instruct} or \textsc{Chat} suffix.
The prompt settings and datasets follow the guidelines from the Chain-of-Thought~\citep{wei2022chain, wang2022self, diao2023active,fu2023hub,pan-etal-2024-plum}, and API model access is provided through the official website. We assessed all open-source models using the vLLM framework~\citep{kwon2023vllm}, which offers efficient KV cache memory management and a Flash attention~\citep{flash-attn,flash-attn2} backend. All open source models run on RTX4090 and RTX A6000 servers.

\subsection{Task configurations}
\label{sec:task_config}
\method{} generates results for three datasets, designated as \textit{\method{}-MMLU}, \textit{\method{}-GSM8K}, and \textit{\method{}-CSQA}. Table \ref{tab:task_config} details the configurations for the \method{} experiments. In this context, $K$ represents the number of questions that the LLM must answer in a single response, while $T$ denotes the number of iterations, also known as query times. The total number of questions addressed is calculated using the formula $K \times T$. To better compare the long-context generation capabilities between API accessed models and open source models, the maximum output length is uniformly set at 4096 tokens in the main experiments. For \textit{\method{}-MMLU}, the $T$ value is considered based on the number of categories. Categories with excessively long input prompts are excluded.
In our main experiment, we arrange the questions in ascending order based on their length, setting the order within a single query from the shortest to the longest length. A detailed ablation study of this variant is discussed in Section \ref{sec:ablation}.

\begin{table}[!h]
\centering
\begin{sc}
\resizebox{\linewidth}{!}{%
\begin{tabular}{l|rccccc}
\toprule
\multirow{3}{*}{Model} & \multicolumn{6}{c}{\method{}}  \\
& \multicolumn{2}{c}{GSM8K} & \multicolumn{2}{c}{MMLU}  & \multicolumn{2}{c}{CSQA} \\
                    & $K$ & $T$ & $K$ & $T$ & $K$ & $T$ \\ \midrule
GPT-3.5-Turbo       & 35  & 20   & 40  & 55  & 80  & 20   \\
GPT-4o              & 35  & 20   & 40  & 55  & 80  & 20   \\
Gemini-1.5-Flash    & 35  & 20   & 40  & 55  & 80  & 20   \\
Claude-3-Haiku      & 30  & 20   & 40  & 55  & 80  & 20   \\ \midrule
LLaMA-3-8B-Instruct & 30  & 20   & 30  & 52  & 40  & 20   \\ 
LLaMA-3-70B-Instruct & 30  & 20   & 30  & 52  & 40  & 20   \\ 
Qwen2-7B-Instruct & 30  & 20   & 30  & 52  & 40  & 20   \\ 
Qwen2-54B-A14B-Instruct & 30  & 20   & 30  & 52  & 40  & 20   \\ 
Qwen2-72B-Instruct & 30  & 20   & 30  & 52  & 40  & 20   \\ 
ChatGLM4-9B-Chat & 30  & 20   & 30  & 52  & 40  & 20   \\ 
DeepSeek-v2-Chat & 30  & 20   & 30  & 52  & 40  & 20   \\ 
\bottomrule
\end{tabular}%
}
\end{sc}
\caption{Configuration details for the \method{} experiment. The table shows the number of questions in one query ($K$) and the number of iteration times ($T$).}
\label{tab:task_config}
\end{table}

\section{Result}
\label{sec:result}

\subsection{API Accessed Models}
Table \ref{tab:longgen_performance_API} displays the performance of various models on the GSM8K and MMLU datasets under two scenarios: Baseline and \method{}. The Delta column shows the change in performance when applying \method{} relative to the Baseline, with negative values indicated by a downward triangle symbol ($\nabla$) signifying performance degradation. The results demonstrate that all models undergo a performance degradation when evaluated under the \method{} conditions. Notably, GPT-3.5-Turbo and Claude-3-Haiku exhibit the largest Delta on both \textit{\method{}-MMLU} and \textit{\method{}-GSM8K}, indicating significant challenges in managing long-context generation. Conversely, the Gemini-1.5-Flash model exhibits the smallest performance degradation, suggesting greater robustness and enhanced consistency in handling long-context scenarios.

\begin{table}[h]
\centering
\begin{sc}
\begin{subtable}[t]{1\linewidth}
\centering
\resizebox{1\linewidth}{!}{
\begin{tabular}{l|ccr}
\toprule
\multirow{2}{*}{Model} & \multicolumn{3}{c}{GSM8K (\%)}   \\
                       & Baseline $\uparrow$ & \method{}$\uparrow$ & Delta$\Delta$       \\ \midrule
GPT-3.5-Turbo          & 75.1     & \cellcolor{red!20}55.3    & -19.8$\nabla$ \\
GPT-4o                 & 91.1     & \cellcolor{red!20}75.6    & -15.5$\nabla$ \\
Gemini-1.5-Flash       & 86.2     & \cellcolor{red!20}85.0    & -1.2$\nabla$  \\
Claude-3-Haiku         & 76.6     & \cellcolor{red!20}55.3    & -21.3$\nabla$ \\ \bottomrule
\end{tabular}
}
\caption{Performance on GSM8K dataset}
\end{subtable}

\vspace{0.5cm}

\begin{subtable}[t]{1\linewidth}
\centering
\resizebox{1\linewidth}{!}{
\begin{tabular}{l|ccr}
\toprule
\multirow{2}{*}{Model} & \multicolumn{3}{c}{MMLU (\%)}   \\
                       & Baseline$\uparrow$ & \method{}$\uparrow$ & Delta $\Delta$       \\ \midrule
GPT-3.5-Turbo          & 70.0     & \cellcolor{red!20}56.3    & -13.7$\nabla$ \\
GPT-4o                 & 88.7     & \cellcolor{red!20}79.7    & -9.0$\nabla$  \\
Gemini-1.5-Flash       & 79.0     & \cellcolor{red!20}74.7    & -4.3$\nabla$  \\
Claude-3-Haiku         & 75.0     & \cellcolor{red!20}48.6    & -26.4$\nabla$ \\ \bottomrule
\end{tabular}
}
\caption{Performance on MMLU dataset}
\end{subtable}
\end{sc}
\caption{Comparison of baseline and LongGen performance on GSM8K and MMLU datasets with API accessed models. }
\label{tab:longgen_performance_API}
\end{table}

\begin{figure}[!h]
\vskip 0.2in
\begin{center}
\includegraphics[width=1\linewidth,trim=0 0 0 0]{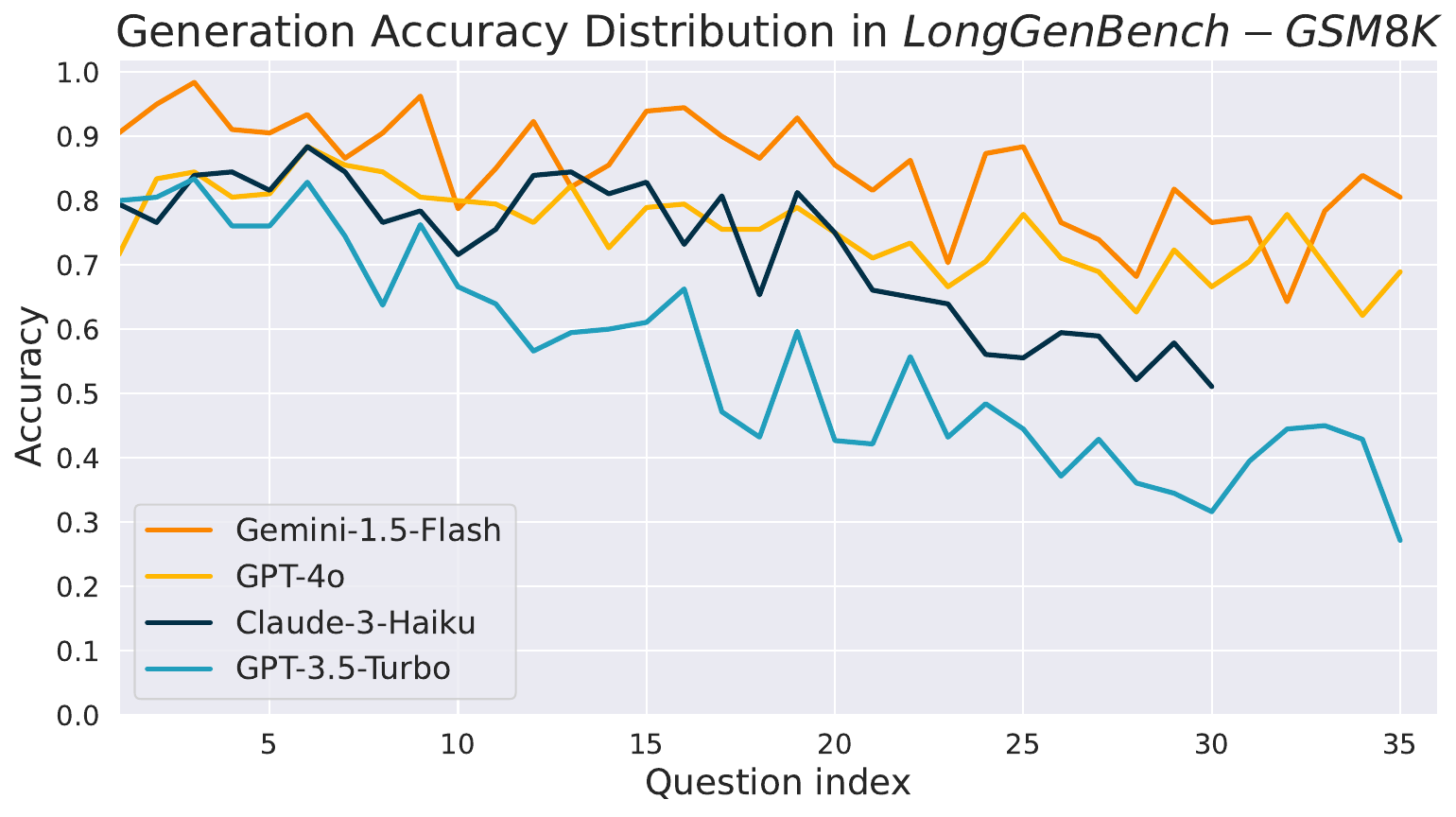}

\caption{Generation accuracy distribution of API accessed models in \textit{\method{}-GSM8K}.}
\label{fig:api_gsm8k}
\end{center}
\end{figure}

Figure \ref{fig:api_gsm8k} shows the accuracy distribution of API accessed models in \textit{\method{}-GSM8K}. The x-axis represents the question index within a single long-text response, with the maximum index being $K$. The y-axis indicates the accuracy of the model's responses to these questions. The analysis reveals how the accuracy varies across different questions for models like GPT-3.5-Turbo, GPT-4o, Gemini-1.5-Flash, and Claude-3-Haiku when they are required to generate answers for $K$ questions simultaneously. The results indicate that all models experience a decline in accuracy as the question index increases. Notably, GPT-3.5-Turbo and Claude-3-Haiku show a more significant decline, suggesting that these models struggle more with maintaining high accuracy over longer sequences of questions. In contrast, Gemini-1.5-Flash maintains relatively higher accuracy, indicating better robustness in handling long-text generation tasks.

\subsection{Open Source Models}

The results presented in Table \ref{tab:longgen_performance_open} provide a comparative analysis of the performance of various open-source models on the GSM8K and MMLU datasets under baseline conditions and using the \method{} approach. Several key observations can be made from these results:

\begin{table}[h]
\centering
\begin{sc}
\begin{subtable}[t]{1\linewidth}
\centering
\resizebox{1\linewidth}{!}{
\begin{tabular}{l|ccr}
\toprule
\multirow{2}{*}{Model} & \multicolumn{3}{c}{GSM8K (\%)}   \\
                       & Baseline$\uparrow$ & \method{}$\uparrow$ & Delta$\Delta$       \\ \midrule
LLaMA-3-8B-Instruct
& 79.6
& \cellcolor{red!20} 32.5
& -47.1$\nabla$ \\

LLaMA-3-70B-Instruct      
& 93.0 
& \cellcolor{red!20} 83.2
& -9.8$\nabla$ \\

Qwen2-7B-Instruct   
& 82.3   
& \cellcolor{red!20}  63.9
& -18.4$\nabla$  \\

Qwen2-57B-A14B-Instruct
& 79.6  
& \cellcolor{red!20}  71.2
& -8.4$\nabla$ \\ 

Qwen2-72B-Instruct 
& 91.1   
& \cellcolor{red!20} 85.7
& -5.4$\nabla$  \\

ChatGLM4-9B-Chat 
& 79.6   
& \cellcolor{red!20}  68.8
& -10.8$\nabla$  \\

DeepSeek-v2-Chat  
& 92.2 
& \cellcolor{red!20} 86.5
& -5.7$\nabla$  \\ 
\bottomrule
\end{tabular}
}
\caption{Performance on GSM8K dataset}
\end{subtable}

\vspace{0.5cm}

\begin{subtable}[t]{1\linewidth}
\centering
\resizebox{1\linewidth}{!}{
\begin{tabular}{l|ccr}
\toprule
\multirow{2}{*}{Model} & \multicolumn{3}{c}{MMLU (\%)}   \\
                       & Baseline$\uparrow$ & \method{}$\uparrow$ & Delta $\Delta$       \\ \midrule
LLaMA-3-8B-Instruct     
&  68.4   
& \cellcolor{red!20}  50.4
& -18.0$\nabla$ \\

LLaMA-3-70B-Instruct  
&   82.0   
& \cellcolor{red!20} 71.2
& -10.8$\nabla$ \\

Qwen2-7B-Instruct
&  70.5  
& \cellcolor{red!20} 59.4
& -11.1$\nabla$  \\

Qwen2-57B-A14B-Instruct  
&  75.4  
& \cellcolor{red!20} 66.7
& -8.7$\nabla$ \\ 

Qwen2-72B-Instruct
&  82.3 
& \cellcolor{red!20} 75.8
& -6.5$\nabla$  \\

ChatGLM4-9B-Chat 
& 72.4     
& \cellcolor{red!20}  63.0
& -9.4$\nabla$  \\

DeepSeek-v2-Chat 
& 77.8  
& \cellcolor{red!20}  72.0
& -5.8$\nabla$  \\
\bottomrule
\end{tabular}
}
\caption{Performance on MMLU dataset}
\end{subtable}
\end{sc}
\caption{Comparison of baseline and LongGen performance on GSM8K and MMLU datasets with open source models.}
\label{tab:longgen_performance_open}
\end{table}
\paragraph{Correlation Between Baseline and \method{}:}There appears to be a general correlation between the baseline performance and the degree of performance degradation observed with the \method{} approach. Models with higher baseline performance tend to exhibit smaller performance drops. For example, Qwen2-72B-Instruct and DeepSeek-v2-Chat models, which have high baseline scores, show relatively small Delta values across both datasets. However, there are exceptions, such as LLaMA-3-70B-Instruct, which despite its high baseline performance, exhibits a significant performance drop on both datasets. Additionally, LLaMA-3-8B-Instruct, Qwen2-57B-Instruct, and ChatGLM4-9B-Chat models have the same baseline score on GSM8K, yet their Delta values differ substantially (47.1\%, 8.4\%, and 10.8\%, respectively).

\paragraph{Impact of Model Size on Performance Degradation:} Observing models within the same series but with different sizes, such as the LLaMA-3 series and the Qwen2 series, reveals a trend where larger models generally exhibit smaller Delta values. This suggests that increasing model size can mitigate performance degradation in long-context generation tasks. For instance, within the LLaMA-3 series, LLaMA-3-70B-Instruct shows a much smaller Delta compared to LLaMA-3-8B-Instruct across both datasets.

\paragraph{Variation Among Different Model Architectures:} Different model architectures demonstrate varying trends in performance degradation. For models within the $7\sim9B$ parameter range, such as LLaMA-3-8B-Instruct, Qwen2-7B-Instruct, and ChatGLM4-9B-Chat, there are notable differences in Delta values despite similar baseline performances. For example, LLaMA-3-8B-Instruct has a Delta of 47.1\% on GSM8K, while ChatGLM4-9B-Chat has a Delta of only 10.8\%, indicating significant variation in how different architectures handle long-context generation tasks.

\paragraph{Consistency Across Tasks for Individual Models:} Individual models exhibit consistent trends in performance degradation across different \method{} tasks. For instance, LLaMA-3-8B-Instruct consistently shows the largest Delta values on both datasets, indicating a significant drop in performance when generating long-context responses. Conversely, Qwen2-72B-Instruct and DeepSeek-v2-Chat consistently show minimal Delta values, suggesting better resilience in long-context tasks.

These findings underscore the importance of considering both model architecture and size when evaluating the performance of LLMs in long-context generation tasks. The \method{} approach effectively highlights the varying capabilities of different models to maintain accuracy over extended text generation, providing valuable insights for further model development and optimization.

\begin{figure}[!h]
\vskip 0.2in
\begin{center}
\includegraphics[width=1\linewidth,trim=0 0 0 0]{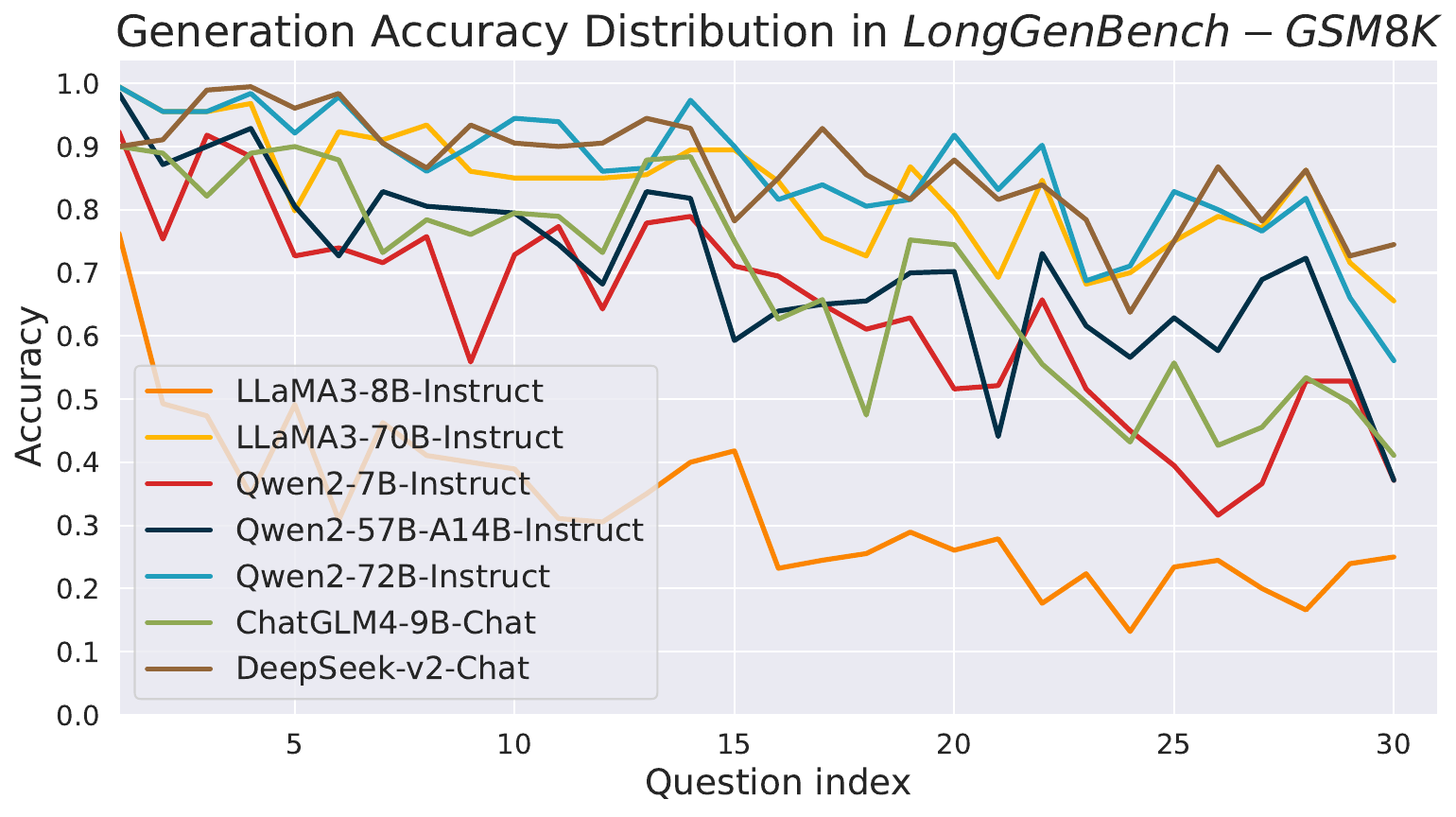}

\caption{Generation accuracy distribution of open source models in \textit{\method{}-GSM8K}. }
\label{fig:open_gsm8k}
\end{center}
\end{figure}

Figure \ref{fig:open_gsm8k} illustrates the accuracy distribution of open-source models in \textit{\method{}-GSM8K}. The results indicate that all models exhibit a decline in accuracy as the question index increases. Notably, LLaMA-3-8B-Instruct experiences more significant performance degradation, suggesting that this model struggles more with maintaining high accuracy in long-context generation tasks.

\subsection{Length Distribution}
\begin{figure}[h]
    \centering
    \includegraphics[width=\linewidth]{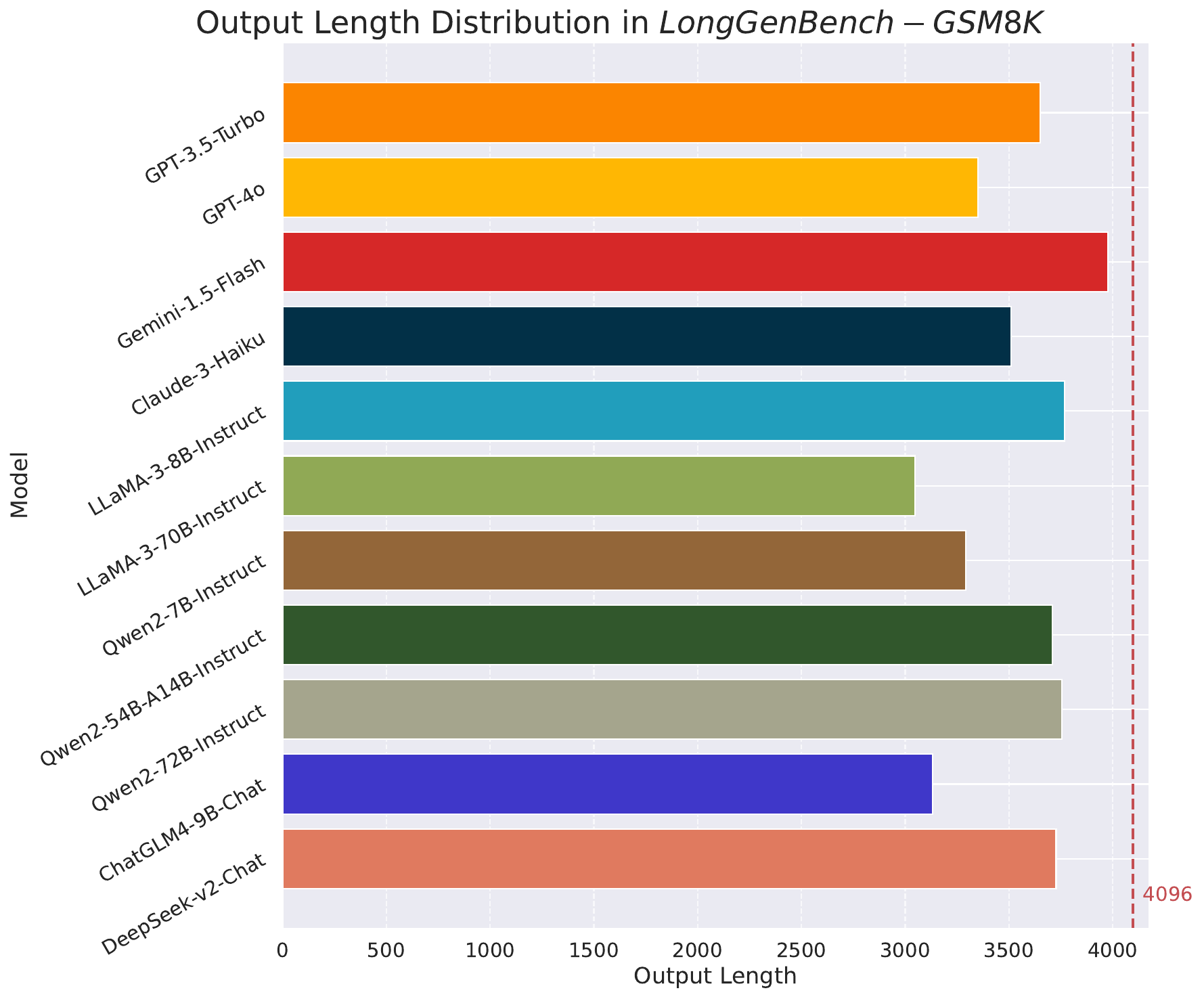}
    \caption{Output Length Distribution in \textit{LongGenBench-GSM8k}.}
    \label{fig:output_length_distribution}
\end{figure}

Figure \ref{fig:output_length_distribution} illustrates the output length distribution for various models in the \textit{\method{}-GSM8K} task, with experimental configurations as detailed in Table \ref{tab:task_config}. Most models produce output lengths close to or exceeding 3500 characters, although none exceed the 4096-character limit. This data demonstrates that \method{} effectively facilitates long-context generation in LLMs.

\section{Ablation Studies}
\label{sec:ablation}
\paragraph{Hyperparameters of \method{}}
To gain deeper insights into \method{}, we conducted an ablation study focusing on two critical hyperparameters: reconstructive processing and the ordering of $K$ questions within a single query. For reconstructive processing, we compared the baseline format with the \method{} format. The baseline format follows the CoT~\citep{wei2022chain} setting, featuring eight question-answer pairs in sequence. In contrast, the \method{} format presents eight questions in advance followed by the corresponding eight answers, with the $K$ questions being addressed subsequently. Regarding the order of $K$ questions, we evaluated three sequences: the original order from the dataset, ascending order from shortest to longest, and descending order from longest to shortest. This ablation study was performed using GPT-3.5-Turbo and Gemini-1.5-Flash on the GSM8k dataset, with the number of iterations set at $T=20$. Table \ref{tab:app_hyperparameters} presents the results of this hyperparameter ablation study, demonstrating that both hyperparameters are crucial for \method{} to effectively evaluate the long-context generation capabilities of LLMs.

\begin{table}[!h]
\centering
\resizebox{\linewidth}{!}{
\begin{tabular}{l|rr|c}
\toprule
\multirow{2}{*}{Model} & \multicolumn{1}{c}{\multirow{2}{*}{Format}} & \multicolumn{1}{c|}{\multirow{2}{*}{Order}} & \multicolumn{1}{c}{\textit{\method{}}} \\
& & & \textit{GSM8K} \\ \midrule
\multirow{4}{*}{\textsc{GPT-3.5-Turbo}} & Baseline & Ascending & 53.2 \\
& \method{} & Ascending & \textbf{55.3} \\
& \method{} & Descending & 51.3 \\
& \method{} & Normal & 47.3 \\ \midrule
\multirow{4}{*}{\textsc{Gemini-1.5-Flash}} & Baseline & Ascending & 82.8 \\
& \method{} & Ascending & \textbf{85.0} \\
& \method{} & Descending & 84.3 \\
& \method{} & Normal & \textbf{85.0} \\ \bottomrule
\end{tabular}%
}
\caption{Performance comparison of different hyperparameter settings.}
\label{tab:app_hyperparameters}
\end{table}

\paragraph{Evaluating Long Input Comprehension}
To address the potential concern that performance degradation in \method{} may be due to the model's inability to comprehend long inputs rather than its ability to generate long outputs, we conducted additional experiments. Specifically, we designed a set of experiments where the model is provided with a long input containing multiple questions but is required to answer only one specified question at a time. This "long input + short output" setting helps isolate the model's comprehension ability from its generation capacity.

For this experiment, we again used GPT-3.5-Turbo and Gemini-1.5-Flash on the GSM8k dataset. We provided the models with a long input sequence of $K$ questions but instructed them to respond to only one randomly selected question per query. This setup was repeated for $T=20$ iterations to ensure robust evaluation.The results, shown in Table \ref{tab:app_long_input_short_output}, indicate that both models maintain high accuracy when required to produce short outputs from long inputs. This supports the hypothesis that the primary challenge in \method{} lies in the generation of long outputs rather than comprehension of long inputs.

\begin{table}[!h]
\centering
\resizebox{\linewidth}{!}{
\begin{tabular}{l|c|c|c}
\toprule
\multirow{2}{*}{Model} & Long Input + & Long Input +  & Performance \\
 & Short Output & Long Output & Drop \\ 
\midrule
GPT-3.5-Turbo & 74.3 & 55.3 & -19.0 \\
Gemini-1.5-Flash & 86.1 & 85.0 & -1.1 \\
\bottomrule
\end{tabular}%
}
\caption{Comparison of model performance in long input + short output versus long input + long output settings.}
\label{tab:app_long_input_short_output}
\end{table}

\section{Conclusion}
In this study, we introduced \method{}, an effective framework designed to evaluate the long-context generation capabilities of language models (LLMs) across multiple datasets. Our experiments included both API accessed and open source models, offering a comprehensive comparison of their performance in long-context generation tasks. The results indicate a correlation between baseline performance and \method{} performance, with higher baseline models generally exhibiting smaller declines. Additionally, model size and architecture significantly influence resilience, with larger models and specific architectures demonstrating greater robustness and consistent trends across different \method{} tasks. These findings highlight the importance of considering both model architecture and size when evaluating LLMs in long-context generation tasks. The \method{} framework effectively showcases the varying capabilities of different models, providing valuable insights for further model development and optimization.


\section{Limitations}
Our study has several limitations. Firstly, the experiments were conducted on a limited set of models and datasets, which may not fully represent the diversity of available LLMs and tasks. Secondly, we did not explore experiments with larger $K$ values due to constraints on the maximum output tokens imposed by API accessed models. Lastly, we did not include experiments with long-context techniques, which may help mitigate the observed performance degradation. These limitations suggest that further research is needed to generalize our findings across a broader range of models and more extended context scenarios.

\section*{Acknowledgments}
This work was supported by the National Natural Science Foundation of China under Grant No. 62272122, the Guangzhou Municipal Joint Funding Project with Universities and Enterprises under Grant No. 2024A03J0616, the Hong Kong RIF grant under Grant No. R6021-20, and Hong Kong CRF grants under Grant No. C2004-21G and C7004-22G, the Guangdong Provincial Department of Education Project (Grant No.2024KQNCX028), the Scientific Research Projects for the Higher-educational Institutions (Grant No.2024312096), Education Bureau of Guangzhou Municipality, the Guangzhou-HKUST(GZ) Joint Funding Program (Grant No.SL2024A03J01201), Education Bureau of Guangzhou Municipality, the China Association for Science and Technology (Grant No.XMSB20240711064).
\bibliography{acl_latex}

\clearpage
\appendix

\section{Experiment Setup and Hyperparameters}
\label{sec:app_exp}

\subsection{Baseline Setting}
The baseline scores referenced in section \ref{sec:result} are derived from official reports to ensure the use of the highest available baseline scores.
\subsection{Dataset}
The statistics of the datasets used in our study are reported in Table~\ref{tab:dataset_statistic}.

\begin{table}[!h]
\centering
\small
\vskip 0.15in
\begin{sc}
    \resizebox{\linewidth}{!}{
    \begin{tabular}{l|rr}
        \toprule
        Dataset  & \# Train & \# Test \\ \midrule
        
        GSM8K~\citep{cobbe2021training}    
        & 7,473     
        & 1,319   \\
        
        MMLU~\citep{hendrycks2020measuring} 
        & - 
        & 14,079 \\

        CSQA*~\citep{talmor2018commonsenseqa} 
        & 9,741
        & 1,221 \\
        
        \bottomrule
        \end{tabular}
    }
\end{sc}
\caption{The statistics of datasets. \textsc{\# Train} and \textsc{\# Test} denote the number of training and test samples respectively. *: CSQA do not have publicly available test set labels, so we simply follow the setting by~\citep{wei2022chain} to evaluate the performance of the development set.}
\label{tab:dataset_statistic}
\end{table}

Table \ref{tab:dataset_num_statistic} below compares the number of instances in LongGenBench with other long-text benchmarks, demonstrating that LongGenBench has a comparable data size in the long-context benchmark field.

\begin{table}[!h]
\centering
\small
\vskip 0.15in
\begin{sc}
    \resizebox{\linewidth}{!}{
    \begin{tabular}{l|r}
        \toprule
        Dataset  &  \# Test \\ \midrule
        \method{} 
        & $16K$ \\
        
        LongBench~\citep{bai2023longbench}    
        & $5K$  \\
        
        $\infty$-Bench~\citep{zhang2024infty} 
        & $5K$ \\

        ZeroSCROLLS~\citep{shaham2023zeroscrolls} 
        & $4K$ \\

        L-Eval~\citep{an2023eval}
        & $2K$ \\
        \bottomrule
        \end{tabular}
    }
\end{sc}
\caption{Comparison of the number of test instances in LongGenBench with other long-text benchmarks, demonstrating that LongGenBench has a comparable data size in the long-context benchmark field.}
\label{tab:dataset_num_statistic}
\end{table}

\subsection{\method{} Format}
\label{sec:app_format}
Table \ref{tab:model_template} compares the baseline method with the \method{} approach in terms of chat templates used for model interactions. In the baseline method, the system prompt is followed by a series of chain-of-thought (CoT) questions and answers, ending with a real question. In contrast, the \method{} approach involves concatenating multiple Chain of Thought (CoT) questions and answers, followed by several real questions, to prompt the model for long-context responses. This method helps evaluate the model's ability to generate coherent and accurate long-context answers across a series of related questions. Additionally, it is easily adaptable to existing benchmarks, allowing for more comprehensive assessments of long-context generation capabilities.   

Table \ref{tab:system_prompt} presents the system prompt for \method{}, which is designed to be straightforward, guiding the LLM to answer each question sequentially.

\begin{table*}[t]
\centering
\footnotesize
\begin{tabular}{p{15cm}}
\toprule
 \multicolumn{1}{c}{\method{} System Prompt Exemplars}\\
\midrule
Answer each question step by step, adhering to the format shown in the examples provided. Start each response with 'Answer\_' and introduce the final response with 'The answer is'. Do not repeat the question. Ensure that you respond to all the questions presented, regardless of their number.\
\\
\bottomrule
\end{tabular}
\caption{\method{} System Prompt Exemplars }
\label{tab:system_prompt}
\end{table*}

\definecolor{prompt}{HTML}{6085e4}
\definecolor{label_color}{HTML}{f19eac}

\subsection{API Models}

In our experiments, we utilized specific versions of various API accessed models to evaluate their performance on \method{}. Table \ref{tab:api_models} provides the details of the models and their respective versions used in our study.

\begin{table}[!h]
\centering
\resizebox{\linewidth}{!}{%
\begin{tabular}{l|l}
\toprule
\textbf{Model} & \textbf{Version} \\ 
\midrule
GPT-3.5-Turbo & GPT-3.5-Turbo-0125 \\ 
GPT-4o & GPT-4o-2024-05-13 \\ 
Gemini-1.5-Flash & Gemini-1.5-Flash-Preview-0514 \\ 
Claude-3-Haiku & Claude-3-Haiku-20240307 \\ 
\bottomrule
\end{tabular}
}
\caption{Specific versions of API models used in the experiments.}
\label{tab:api_models}
\end{table}

These versions were selected based on their availability and state-of-the-art performance at the time of experimentation. Each model was tested using the \method{} framework to assess their capabilities in handling long-context generation tasks. The results presented in this paper reflect the performance of these specific versions, providing a comprehensive comparison across different models and their configurations.

\section{Additional Experiments}

\subsection{API Accessed Models}
\begin{figure}[!h]
\vskip 0.2in
\begin{center}
\includegraphics[width=1\linewidth,trim=0 0 0 0]{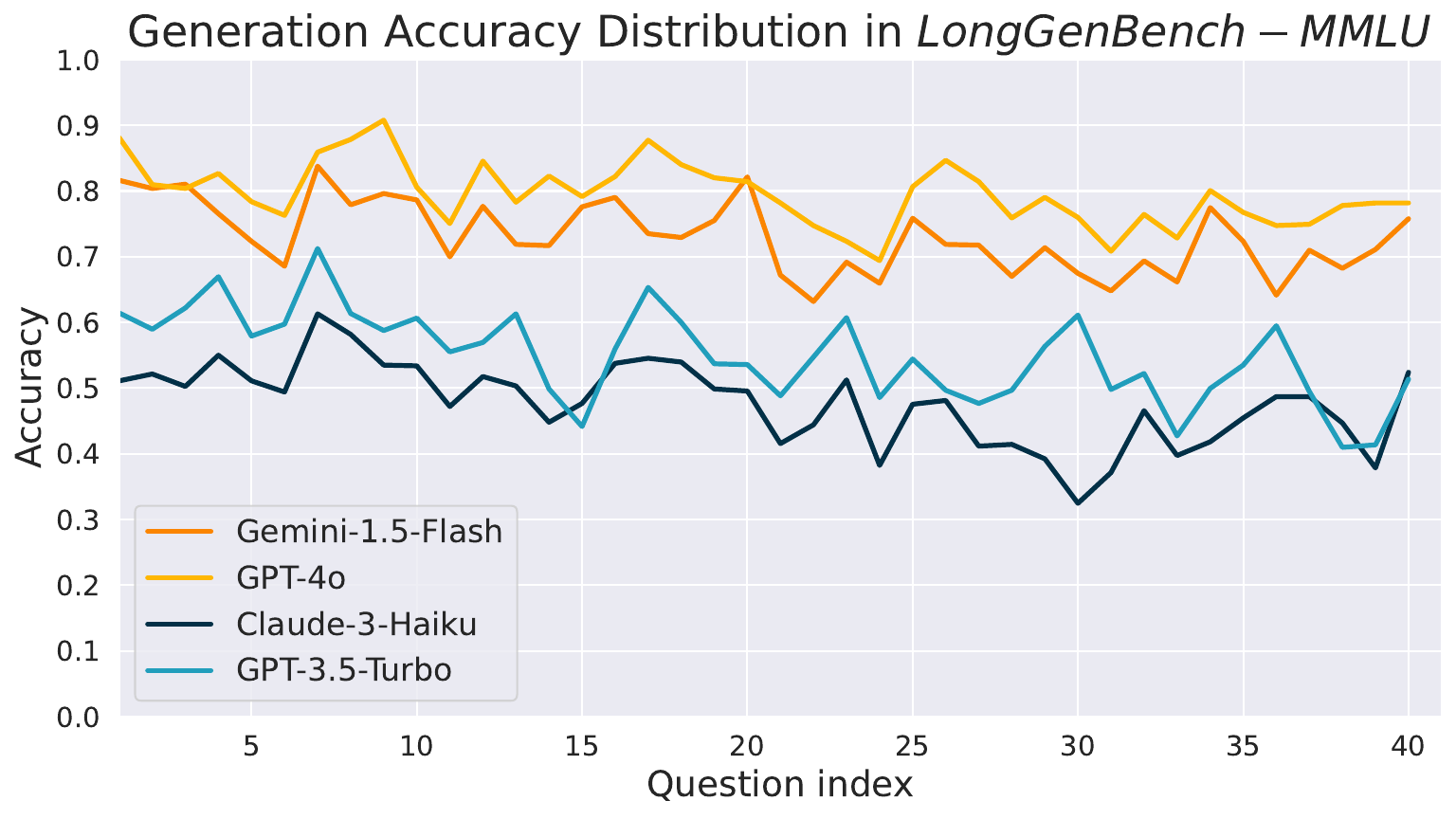}
\includegraphics[width=1\linewidth,trim=0 0 0 0]{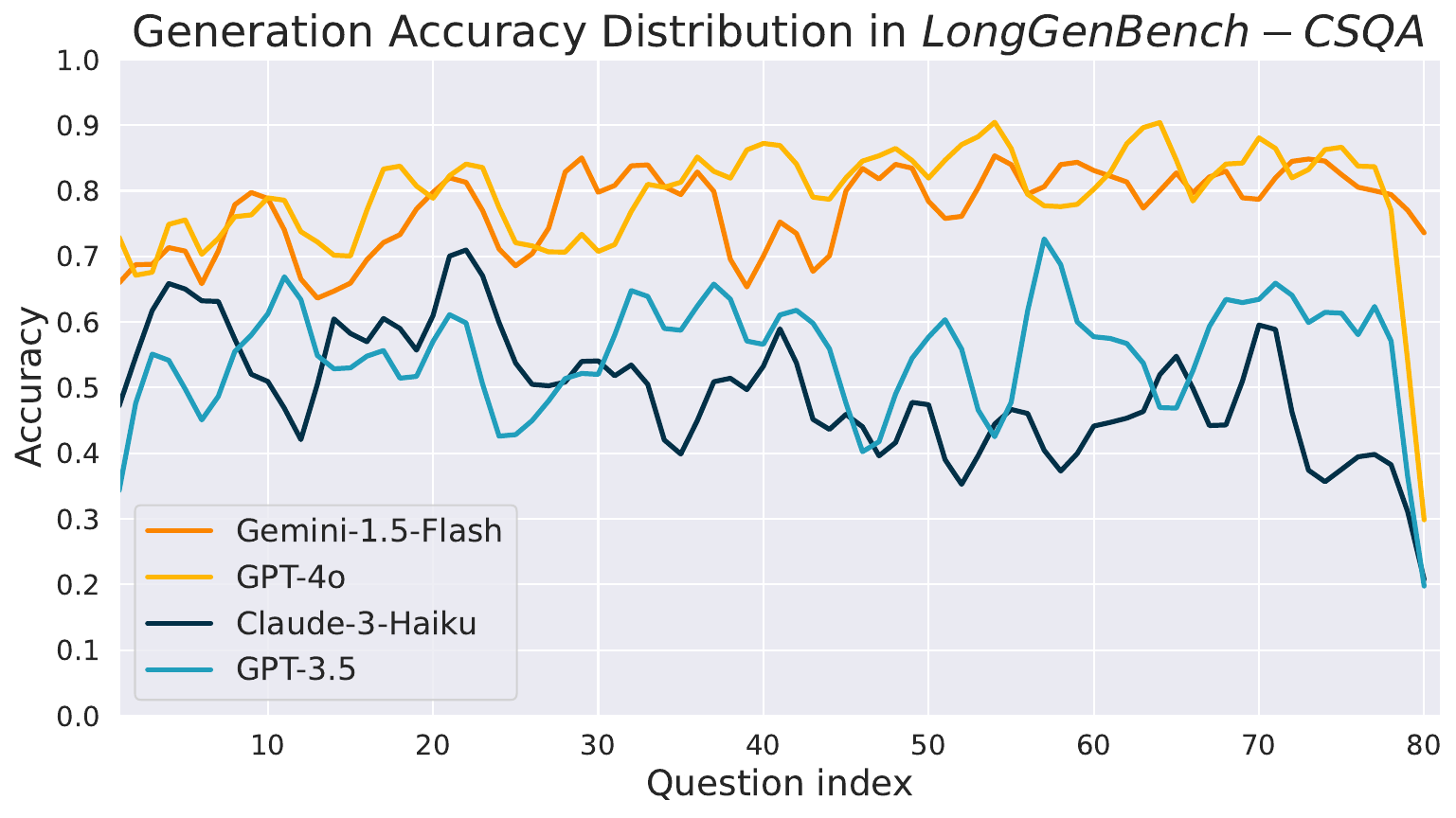}
\caption{Generation accuracy distribution of API accessed models in \textit{\method{}-MMLU} and \textit{\method{}-CSQA}.}
\label{fig:app_api}
\end{center}
\end{figure}

Figure \ref{fig:app_api} and Table \ref{tab:longgen_performance_API_csqa} present the generation accuracy distribution for API accessed models in \textit{\method{}-MMLU} and \textit{\method{}-CSQA}. The x-axis represents the question index within a single long-text response, with the maximum index being $K$. The y-axis indicates the accuracy of the model's responses to these questions. The analysis demonstrates how the accuracy varies across different questions for models like GPT-3.5-Turbo, GPT-4o, Gemini-1.5-Flash, and Claude-3-Haiku when they are required to generate answers for $K$ questions simultaneously. The results reveal that all models experience a decline in accuracy as the question index increases, with GPT-3.5-Turbo and Claude-3-Haiku showing more significant declines. Conversely, Gemini-1.5-Flash and GPT-4o maintains relatively higher accuracy, indicating better robustness in handling long-text generation tasks. Since these models do not provide official results for CSQA, we use our replicated baseline score.

\begin{table}[h]
\centering
\begin{sc}
\centering
\resizebox{1\linewidth}{!}{
\begin{tabular}{l|ccr}
\toprule
\multirow{2}{*}{Model} & \multicolumn{3}{c}{CSQA (\%)}   \\
                       & Baseline$\uparrow$ & \method{}$\uparrow$ & Delta $\Delta$       \\ \midrule
GPT-3.5-Turbo          & 75.57     & \cellcolor{red!20}61.88    & -13.87$\nabla$ \\
GPT-4o                 & 85.75     & \cellcolor{red!20}77.88    & -7.87$\nabla$  \\
Gemini-1.5-Flash       & 83.87     & \cellcolor{red!20}82.25    & -1.62$\nabla$  \\
Claude-3-Haiku         & 66.75     & \cellcolor{red!20}55.25    & -11.50$\nabla$ \\ \bottomrule
\end{tabular}
}
\caption{Comparison of baseline and LongGen performance on CSQA datasets with API models.}
\label{tab:longgen_performance_API_csqa}
\end{sc}
\end{table}

\begin{table*}[h]
    \centering
    \begin{tabular}{p{0.15\linewidth}|p{0.75\linewidth}}
    \toprule
    \textbf{Method} & \textbf{Template} \\ 
    \midrule
    
    Baseline & \textcolor{violet}{\{System Prompt\}} \\
        &\textcolor{prompt}{\{CoT Question\_1\}}\textcolor{orange}{\{CoT Asnwer\_1\}} \\ 
        &$\cdots$ \\
        &\textcolor{prompt}{\{CoT Question\_8\}}\textcolor{orange}{\{CoT Asnwer\_8\}} \\ 
        & \textcolor{label_color}{\{Real Question\}} \\ \midrule
    
    \method{}&  \textcolor{violet}{\{System Prompt\}}  \\
        &\textcolor{prompt}{\{CoT Question\_1\}}$\cdots$\textcolor{prompt}{\{CoT Question\_8\}} \\
        &\textcolor{orange}{\{CoT Asnwer\_1\}}$\cdots$\textcolor{orange}{\{CoT Asnwer\_8\}} \\
        &\textcolor{label_color}{\{Real Question\_1\}}  $\cdots$  \textcolor{label_color}{\{Real Question\_$K$\}} \\ 
    \bottomrule
    \end{tabular}
    \caption{Model chat templates.}
    \label{tab:model_template}
\end{table*}

\begin{figure}[!h]
\vskip 0.2in
\begin{center}
\includegraphics[width=1\linewidth,trim=0 0 0 0]{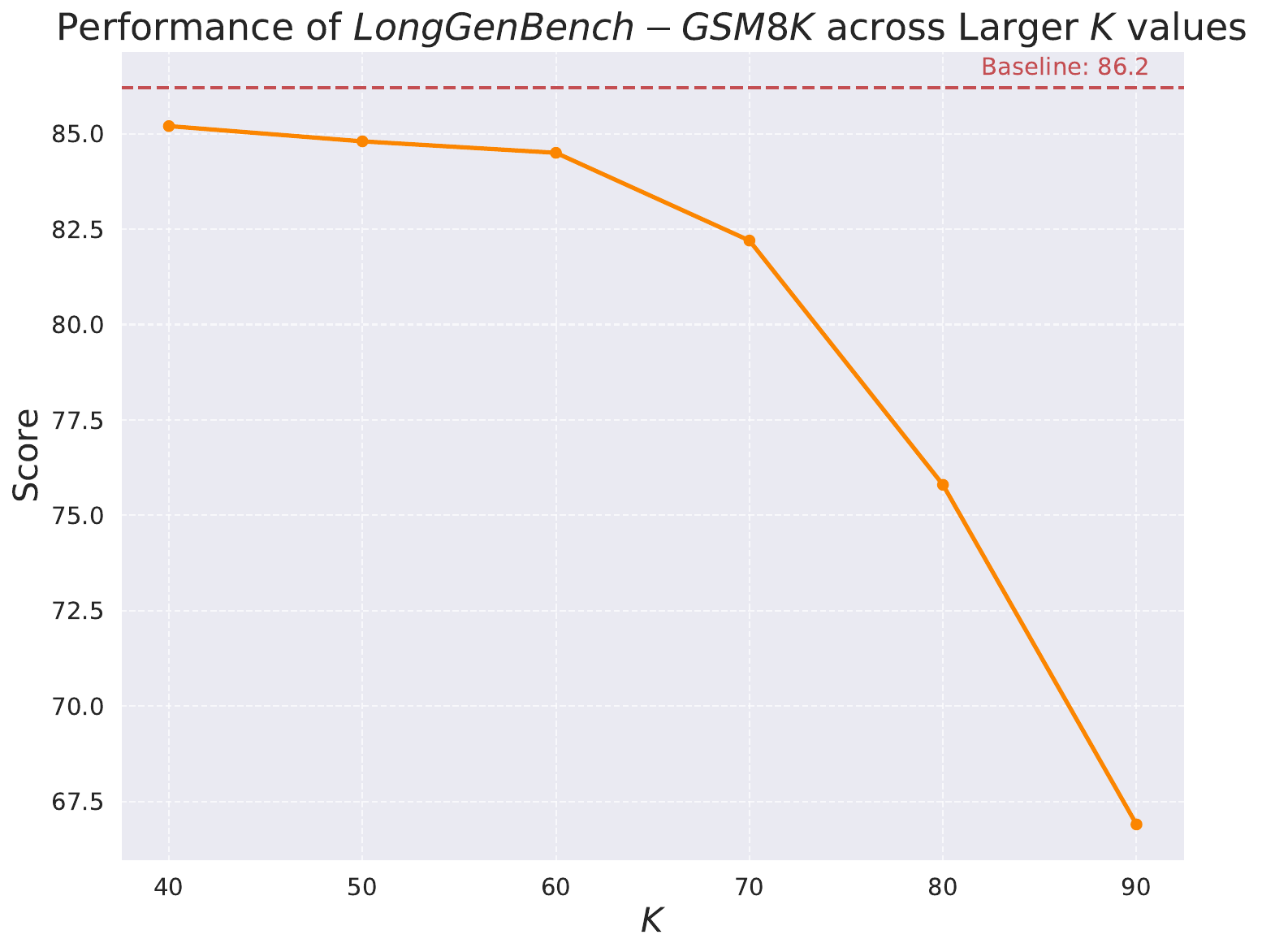}
\caption{Performance of Gemini-1.5-Flash in \textit{\method{}-GSM8K} with larger $K$ values.}
\label{fig:app_larger_K}

\end{center}
\end{figure}

In addition to the main experiments, we conducted an extended evaluation using the Gemini-1.5-Flash model, which supports a longer maximum output length $8K$ tokens. For this extended evaluation, we set the value of $K$ in the range ${40, 50, 60, 70, 80, 90}$ and fixed the number of iterations $T$ at 10. This allowed us to investigate the impact of larger $K$ values on model performance in the \textit{\method{}-GSM8K} experiment. Figure \ref{fig:app_larger_K} shows the performance scores as $K$ increases from 40 to 90, with the number of iterations $T$ set to 10. A red dashed line indicates the baseline score of 86.2 for comparison.

\subsection{Open Source Model}
\begin{figure}[!h]
\vskip 0.2in
\begin{center}
\includegraphics[width=1\linewidth,trim=0 0 0 0]{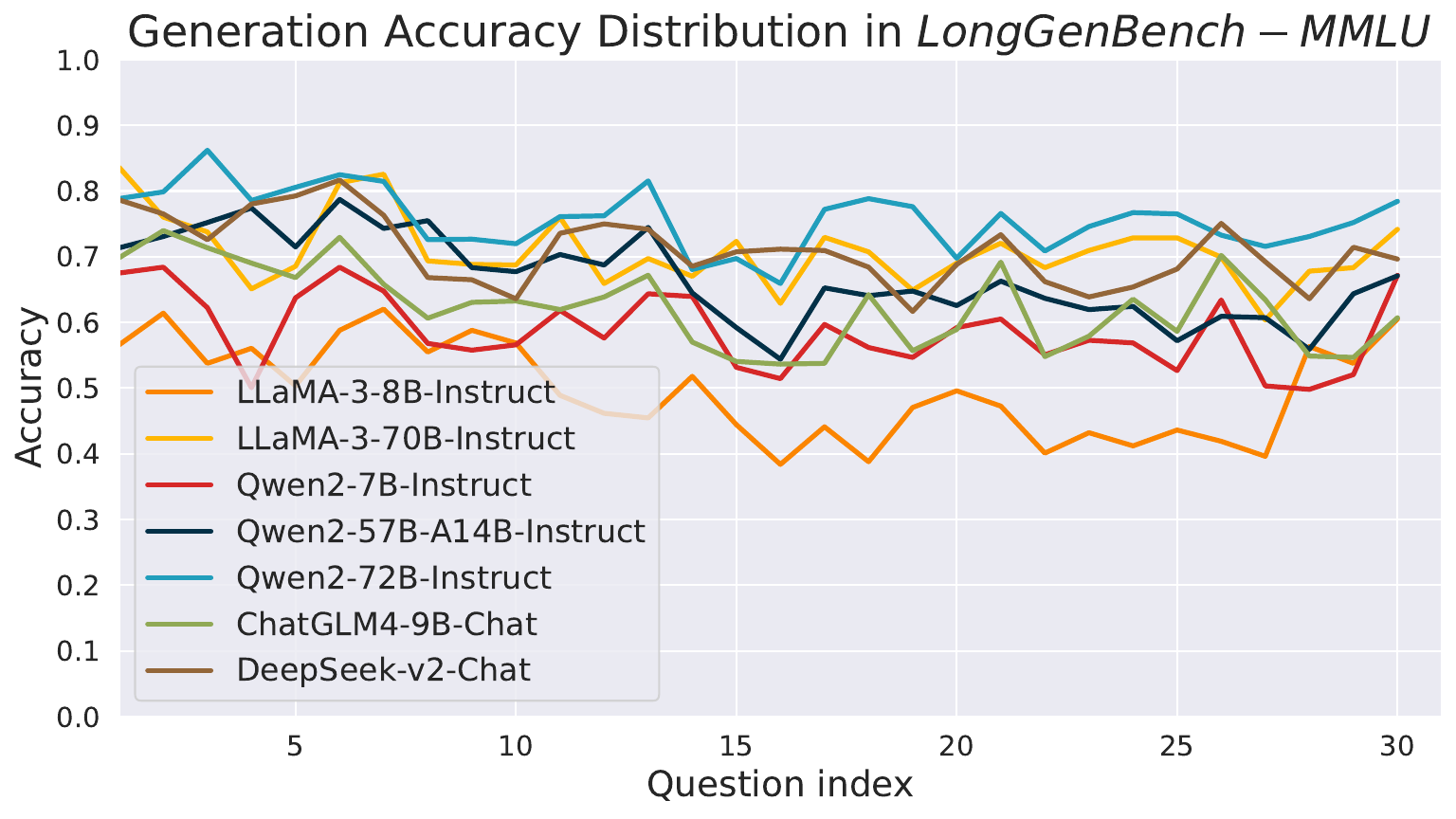}
\includegraphics[width=1\linewidth,trim=0 0 0 0]{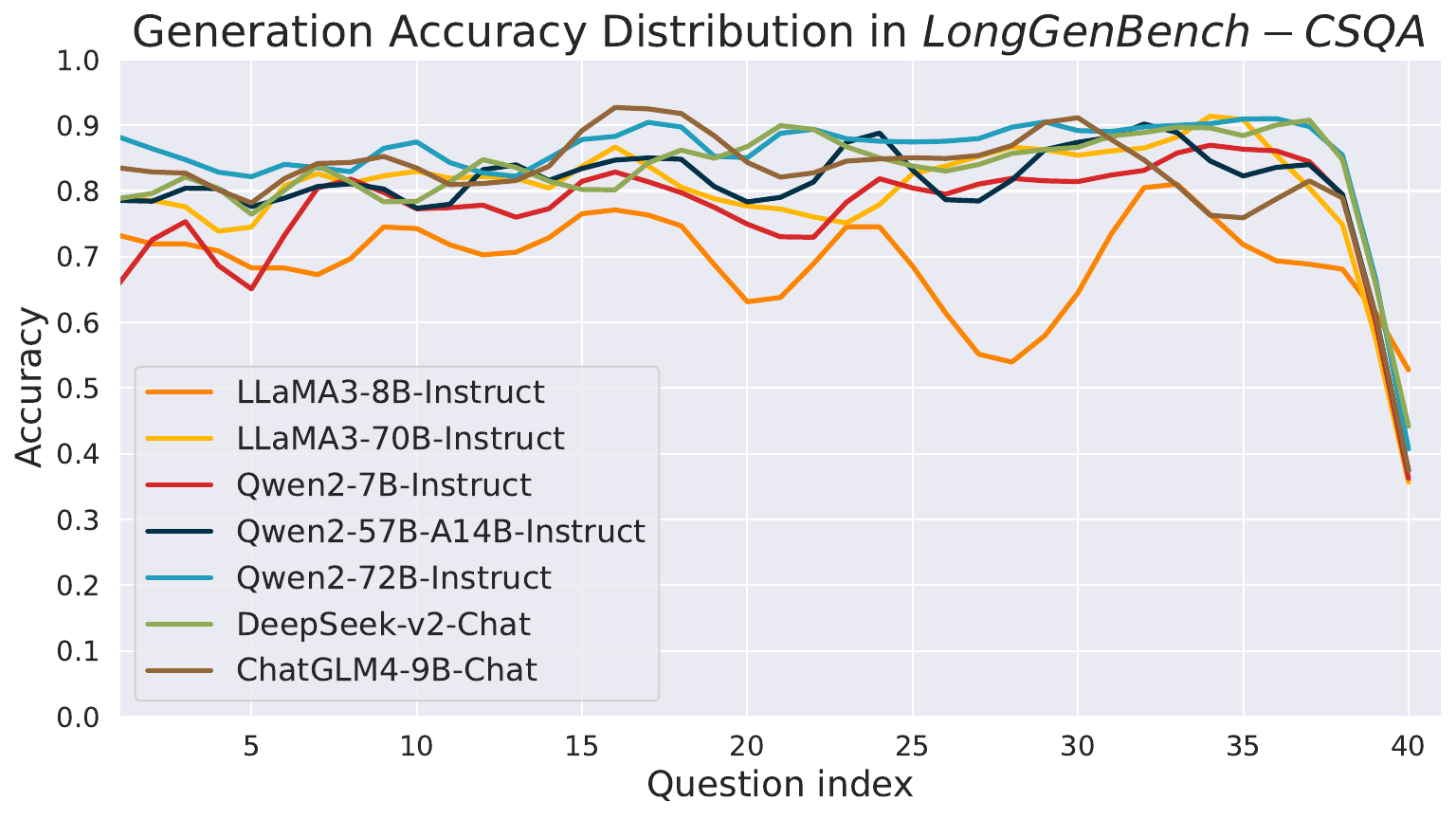}
\caption{Generation accuracy distribution of open source models in \textit{\method{}-MMLU} and \textit{\method{}-CSQA}.}
\label{fig:app_open}
\end{center}
\end{figure}

Figure \ref{fig:app_open} and Table \ref{tab:longgen_performance_open_csqa} illustrates the generation accuracy distribution for open source models in \textit{\method{}-MMLU} and \textit{\method{}-CSQA}. Similar to the API accessed models, the x-axis represents the question index, and the y-axis represents the accuracy of responses. The results indicate that all open source models exhibit a decline in accuracy as the question index increases. Notably, LLaMA-3-8B-Instruct experiences significant performance degradation, suggesting that this model struggles with maintaining high accuracy in long-context generation tasks. In contrast, larger models such as LLaMA-3-70B-Instruct and Qwen2-72B-Instruct demonstrate greater resilience, maintaining higher accuracy across longer sequences of questions. Since these models do not provide official results for CSQA, we use our replicated baseline score.

\begin{table}[h]
\centering
\begin{sc}
\centering
\resizebox{1\linewidth}{!}{
\begin{tabular}{l|ccr}
\toprule
\multirow{2}{*}{Model} & \multicolumn{3}{c}{CSQA (\%)}   \\
                       & Baseline$\uparrow$ & \method{}$\uparrow$ & Delta $\Delta$       \\ \midrule
LLaMA-3-8B-Instruct     
&  73.70   
& \cellcolor{red!20}  69.50
& -4.20$\nabla$ \\

LLaMA-3-70B-Instruct  
&   81.82   
& \cellcolor{red!20} 80.13
& -1.69$\nabla$ \\

Qwen2-7B-Instruct
&  78.13
& \cellcolor{red!20} 77.25
& -0.88$\nabla$  \\

Qwen2-57B-A14B-Instruct  
&  80.26 
& \cellcolor{red!20} 80.75
& +0.49$\Delta$ \\ 

Qwen2-72B-Instruct
&  87.75 
& \cellcolor{red!20} 85.50
& -2.25$\nabla$  \\

ChatGLM4-9B-Chat 
& 85.42     
& \cellcolor{red!20}  82.37
& -3.05$\nabla$  \\

DeepSeek-v2-Chat 
& 84.77 
& \cellcolor{red!20}  82.87
& -1.9$\nabla$  \\
\bottomrule
\end{tabular}
}
\end{sc}
\caption{Comparison of baseline and LongGen performance on CSQA datasets with open source models.}
\label{tab:longgen_performance_open_csqa}
\end{table}

\subsection{Cost}
In this section, we compare the costs between the Needle-In-A-Haystack (NIAH) test and \method{}. For NIAH, costs are calculated for a 128K token test. For \method{}, calculations are based on the same \(K\) and \(T\) values used in the GPT-4o configuration, as specified in Table \ref{tab:task_config}, and applied across the datasets \textit{\method{}-GSM8K}, \textit{\method{}-MMLU}, and \textit{\method{}-CSQA}. The pricing is based on the GPT-4o rates, which are \$5 per million input tokens and \$15 per million output tokens.

As shown in Table \ref{tab:app_cost}, the NIAH test incurs a significantly higher cost compared to \method{}. Specifically, the NIAH test requires 8192K input tokens and 0.1K output tokens, resulting in a total cost of \$40. In contrast, \method{} requires 300K input tokens and 340K output tokens, resulting in a much lower total cost of \$6.6.

\begin{table}[!h]
\centering
\resizebox{\linewidth}{!}{%
\begin{tabular}{l|ccc}
\toprule
Method                 & Input Tokens & Output Tokens &  Price \\ \midrule
NIAH                   &    8192K    &  0.1K &  \$40    \\
\method{}         &   300K     & 340K   &  \$6.6   \\ \bottomrule
\end{tabular}%
}
\caption{Comparison of Costs Between Needle-In-A-Haystack (NIAH) and \method{} Tests}
\label{tab:app_cost}
\end{table}

\section{Exemplars Generated by LLMs Using \method{}}
Tables \ref{tab:example_gsm8k} and \ref{tab:example_mmlu} provide examples of responses generated by LLMs using \method{}, where each model generated $K$ answers in a single long-context response. These examples demonstrate that the models can remember the content and sequence of the $K$ questions and maintain consistency in logical flow during generation.

\begin{table*}[t]
\centering
\footnotesize
\begin{tabular}{p{15cm}}
\toprule
 \multicolumn{1}{c}{Exemplars}\\
\midrule
Answer\_1:\\
Terry eats 2 yogurts a day, so in 30 days, he eats $30 * 2 = 60$ yogurts.
Since yogurts are on sale at 4 for \$5.00, he spends $60 / 4 * \$5.00 = \$75.00$ on yogurt over 30 days.    \emojicross \\
\\
Answer\_2:\\
There are 25 oranges in total, and 20\% of them are unripe, so 20\% of 25 is $0.20 * 25 = 5$.
1 is bad and 2 are sour, so from $25 - (1+2+5) = 25 - 8 = 17$
Therefore, 17 oranges are good. \emojicheck \\ 
\\
Answer\_3:\\
There are 6 clusters with $6 * 20 = 120$ raspberries.
Together with the 67 individual fruit, there are $120 + 67 = 187$ raspberries in total. \emojicheck \\
\\
Answer\_4: \\
Lloyd produces 252 eggs per day, so in a week, he produces 252 * 7 = 1764 eggs.
He sells them for \$2 per dozen, so he sells $1764 / 12 = 147$ dozens per week.
Therefore, his weekly income from eggs is $147 * \$2 = \$294$. \emojicheck \\ 
\\
Answer\_5: \\
If there are twice as many boys as girls in the school and there are 60 girls, then there are $2 * 60 = 120$ boys.
With 5 students to every teacher, there are $60 + 120 = 180$ students.
Therefore, there are 180 / 5 = 36 teachers. \emojicheck \\
\\
\ldots \ldots
\\ \\
Answer\_24: \\
Each school sent 2 teams with 5 players and 2 coaches, so each school sent $2 * (5 + 2) = 14$ people.
With 4 schools, there are therefore $4 * 14 = 56$ people in total. \emojicross \\
\\
Answer\_25: \\
If Dana spends 1/3 of 6 hours running, she runs $1/3 * 6 = 2$ hours.
She travels at a rate of $3 * 2 = 6$ miles per hour.
If she spends 2/3 of 6 hours walking, she walks $2/3 * 6 = 4$ hours.
She travels at $1/4 * 3 = 0.75$ miles per hour, so she walks $4 * 0.75 = 3$ miles.
In total, she travels $2 * 6 + 4 * 3 = 12 + 12 = 24$ miles in 6 hours. \emojicross \\
\\
Answer\_26: \\
The final price of Stephen's groceries is $25\% + \$3.00 + \$4.00 = \$13.00$ more than \$40.00, so the final price is $\$40.00 + \$13.00 = \$53.00$. \emojicross \\
\\ 
Answer\_27:\\
In the first half of the year, Aleena pays $\$140 * 6 = \$840$.
After the discount, in the second half of the year, she pays $10\% * \$140 * 6 = \$84$ less, so she pays $\$840 - \$84 = \$756$.
Thus, the total amount she paid for the streaming service by the end of the year is $\$840 + \$756 = \$1596$. \emojicheck \\
\\
Answer\_28:\\
The truck and driver weigh 3755 pounds, leaving a maximum capacity of $5000 - 3755 = 1245$ pounds.
So up to $1245 / 15 = 83$ boxes can be loaded onto the truck. \emojicheck \\
\\
\ldots \ldots
\\ \\
Answer\_32: \\
Sophia travels 100 miles on 4 gallons of gas, so she uses $100 / 4 = 25$ miles per gallon.
On one tank of 12 gallons, she will be able to drive $25 * 12 = 300$ miles. \emojicheck \\
\\
Answer\_33: \\
For repairing 6 truck tires, the revenue is $6 * \$60 = \$360$.
For repairing 4 car tires, the revenue is $4 * \$40 = \$160$.
The difference in revenue is $\$360 - \$160 = \$200$ more. \emojicross \\ \\
Answer\_34:\\
Meredith spent 4 hours on each blog article, so 5 articles took $5 * 4 = 20$ hours on Monday.
On Tuesday, she wrote 2/5 more articles, which is $2/5 * 5 = 2$, making a total of $5 + 2 = 7$ articles.
Then on Wednesday, she wrote twice the number of articles as on Tuesday, so $7 * 2 = 14$ articles.
All in all, Meredith spent $20 + (7 * 4) + (14 * 4) = 20 + 28 + 56 = 104$ hours writing articles in the three days. \emojicheck \\
\\
Answer\_35:\\
If Polly holds the sapphire gemstone, she can throw the gold javelin three times as far, so she can throw it $400 * 3 = 1200$ feet.
Therefore, holding the sapphire gemstone, she can stand $1200 - 1000 = 200$ feet outside the reach of the dragon's flames and still hit the dragon. \emojicheck \\
\bottomrule
\end{tabular}
\caption{Exemplars for GPT-3.5-Turbo generated response from \textit{\method{}-GSM8K}. }
\label{tab:example_gsm8k}
\end{table*}

\begin{table*}[t]
\centering
\footnotesize
\begin{tabular}{p{15cm}}
\toprule
 \multicolumn{1}{c}{Exemplars}\\
\midrule
Answer\_1:\\
Let's think step by step. Bubblesort has $O(n^2)$ running time in both worst and average cases. Mergesort has $O(n log n)$ running time in both worst and average cases. Heapsort has $O(n log n)$ running time in both worst and average cases. Quicksort has $O(n^2)$ running time in the worst case and $O(n log n)$ running time on average. The answer is (D).  \emojicheck \\
\\
Answer\_2:\\
Let's think step by step. Insertion sort is heavily dependent on the initial ordering of the input. Quicksort is heavily dependent on the initial ordering of the input. Merge sort is least dependent on the initial ordering of the input, because its performance is consistent regardless of the ordering of the input. Selection sort is also heavily dependent on the initial ordering of the input. The answer is (C).\emojicheck \\
\\
Answer\_3:\\
Let's think step by step. We can think of a K-ary tree as being composed of one nonterminal root node, K nonterminal child nodes of the root, $K^2$ nonterminal child nodes of each of those nodes, etc. Thus, a K-ary tree of depth N is composed of:
$1 + K + K^2 + .... + K^N$ nonterminal nodes
and
$(K^N - 1)/(K - 1)$ total nodes.
As the depth of the tree increases, the proportion of nonterminal nodes relative to the total number of nodes approaches:
$(1 + K + K^2 + .... + K^N) / ((K^N - 1)/(K - 1)) =
((K^(N+1) - 1)/(K - 1)) / ((K^N - 1)/(K - 1)) =
(K^(N+1) - 1) / (K^N - 1) =
(K^(N+1) - 1) / (K^N - 1) * (K/K) =
(K^(N+2) - K)/(K^(N+1) - K) =
(K^(N+1) - 1)/(K^N - 1) + 1 =
(1 + K + K^2 + .... + K^(N+1)) / (1 + K + K^2 + .... + K^N)$
which as N increases approaches:
$K/(K - 1)$.
The answer is (B). \emojicross \\
\\
Answer\_4:\\
Let's think step by step. We can see that \{AND, OR\} is not complete because we cannot express NOT. We can see that \{AND, NOT\} and \{NOT, OR\} are both complete. We can also see that \{NAND\} is complete because NAND can express all other boolean operations. The answer is (C). \emojicheck \\ \\
\ldots \ldots \\
\\

Answer\_19:\\
Let's think step by step. The IP protocol is primarily concerned with routing packets through the network. It defines the format of packets and the addressing scheme used to route packets between different networks. The answer is (A). \emojicheck \\
\\
Answer\_20: \\
Let's think step by step. TCP protocol is responsible for reliable delivery of large (multi-packet) messages between machines that are not necessarily directly connected. It provides a reliable connection-oriented service, ensuring that packets arrive in the correct order and are not lost. The answer is (C). \emojicheck \\
\\
Answer\_21: \\
Let's think step by step. A common divisor must divide both integers. The set of common divisors of two integers is a subset of the divisors of each integer. Thus, the set \{-6,-2, -1, 1, 2, 6\} could possibly be the set of all common divisors of two integers. The answer is (A). \emojicross \\
\\
\ldots \ldots \\
\\
Answer\_38: \\
Let's think step by step. If the owner is allowed to delete a file, dangling links may result. This is true, because if the owner deletes the file, the link count in the shared file node will be decremented. However, the links to the file in other users' directories will still point to the shared file node. If the owner deletes the file, these links will become dangling links. The answer is (C). \emojicheck \\
\\
Answer\_39:\\
Let's think step by step. Giving priorities to processes and ordering the wait queues by priority will not deal effectively with the problem of deadlock. This is because deadlock can occur even if processes have priorities and wait queues are ordered by priority. For example, consider a system with two processes, P1 and P2, and two resources, R1 and R2. P1 holds R1 and requests R2, while P2 holds R2 and requests R1. Deadlock will occur even if P1 has a higher priority than P2. The answer is (A). \emojicheck \\
\\

Answer\_40: \\
Let's think step by step. Indirect representation noticeably increases compilation time. This is false, because indirect representation does not noticeably increase compilation time. Direct representation decreases the time needed to access components of a variable. This is true, because direct representation allows for faster access to components of a variable, as the compiler does not need to dereference a pointer to access the components. When the storage size of some private component of a variable changes, indirect representation minimizes the number of recompilations of source modules that must be performed. This is true, because indirect representation allows for changes to the size of private components without requiring recompilation of source modules that use the variable. The answer is (D). \emojicheck\\
\\
\bottomrule
\end{tabular}
\caption{Exemplars for Gemini-1.5-Flash generated response from \textit{\method{}-MMLU}. }
\label{tab:example_mmlu}
\end{table*}

\section{Licenses}
\label{app_licenses}
For the evaluation dataset, all the datasets, including GSM8K~\citep{cobbe2021training}, MMLU~\citep{hendrycks2020measuring}, CSQA~\citep{talmor2018commonsenseqa} are released under MIT license.

\end{document}